\documentclass{article}
\usepackage{stywhispers,amsmath,epsfig}
\usepackage[dvipsnames, table]{xcolor}
\definecolor{myblue}{HTML}{0068F0}
\definecolor{myorange}{HTML}{F08700}
\definecolor{mygreen}{HTML}{00A30B}
\definecolor{myred}{HTML}{F01000}
\definecolor{myyellow}{HTML}{FFF28C}
\usepackage[colorlinks=true, citecolor=myblue, linkcolor=myorange, urlcolor=myblue]{hyperref}
\usepackage{multirow}
\usepackage{makecell}
\usepackage{fixltx2e}
\usepackage{placeins}
\usepackage{pifont}
\def\hlinewd#1{%
\noalign{\ifnum0=`}\fi\hrule \@height #1 %
\futurelet\reserved@a\@xhline} 
\usepackage{amssymb}
\usepackage{moresize}
\usepackage{dirtytalk}
\usepackage{tikz}
\usepackage{hhline}

\title{nnAEU: A Simple and Efficient CNN-Autoencoder for Hyperspectral Unmixing}
\name{Edgard~Dabier, Christophe~Kervazo, Pietro~Gori and Florence~Tupin\thanks{C. Kervazo would like to thank the ANR agency for the grant ANR-25-CE48-1042.}}
\address{LTCI, Télécom Paris, Institut Polytechnique de Paris, Palaiseau, France}

\begin{document}
\maketitle

\begin{abstract}
Deep learning-based hyperspectral unmixing (HSU) has followed a trajectory of increasing architectural complexity, from simple autoencoders to transformers and, most recently, Mamba-based models, each claiming to advance the state of the art. Contrary to this trend, we propose a lightweight CNN-based autoencoder, coined \say{Not New Autoencoder Unmixing} (nnAEU). It is derived from the backbone of UNMamba (2025) through a pruning study that removes the Mamba component entirely. nnAEU uses less than 10 times fewer parameters than UNMamba, and far fewer than other transformer- and GAN-based methods, while outperforming them in accuracy. We provide a complete, minimal, reproducible code, available at \textcolor{myblue}{\textit{https://gitlab.telecom-paris.fr/ring/nnAEU.git}}.
\end{abstract}

\section{Introduction}
\label{sec:intro}

\subsection{Context}

Hyperspectral images (HSIs) capture multiple narrow, adjacent spectral bands, enabling material analysis in remote sensing with applications in agriculture, mineral mapping, and space exploration \cite{Ren_2026}. However, due to physical limitations, HSIs often have low spatial resolution: each pixel covers a large area (\emph{e.g.}, $\sim$30 meters for satellite images) and may contain mixed material spectra. This mixing limits direct interpretation, motivating hyperspectral unmixing (HSU), which decomposes pixels into pure material spectra (endmembers) and their relative abundances. Specifically, this article focuses on blind HSU, aiming to estimate both endmembers and abundances. Despite its limitations \cite{kervazo2021provably}, the Linear Mixture Model (LMM) is widely used for its simplicity. It assumes pixels to be linear combinations of endmembers \cite{Ren_2026}. Mathematically, for an HSI $Y \in \mathbb{R}^{B, N}$ (with $N = H \times W$ pixels and $B$ bands), endmember matrix $E \in \mathbb{R}^{B, C}$, and abundance maps $A \in \mathbb{R}^{C, N}$, the model is:
\vspace{-10pt}
\begin{equation}
    Y = EA + \varepsilon,
\end{equation}

where $\varepsilon \in \mathbb{R}^{B, N}$ is additive noise. The LMM is typically constrained by non-negativity (ANC) and sum-to-one (ASC) conditions for endmembers and abundances.

\subsection{Related work}
\label{sec:sota}

Due to their simplicity, pure pixel retrieval algorithms such as VCA \cite{Nascimento_2005} or N-FINDR \cite{Winter_1999} gained popularity in HSU. However, their reliance on the pure-pixel assumption —rarely satisfied in practice— led, during the last decade, to the dominance of optimisation-based methods with priors such as sparsity \cite{Bioucas_2010} or minimum volume \cite{Heylen_2011}, or even learned priors \cite{kervazo2026unrolled}.

More recently, the application of deep learning to HSU has followed a trajectory of increasing architectural complexity. Early approaches used simple autoencoders for unsupervised endmember and abundance estimation \cite{Palsson_2018}, soon evolving into convolutional variants like CNNAEU \cite{Palsson_2021} and UnDIP \cite{Rasti_2022} to better exploit spatial context. Attention mechanisms and transformers, such as DeepTrans \cite{Ghosh_2022} and A2SAN \cite{Tao_2024}, were later introduced to model long-range spectral-spatial dependencies, while adversarial schemes like SSST-GAN \cite{Zhang_2026} enabled blind HSU by learning data distributions. There have also been attempts at using Hyperspectral Foundation Models for HSU \cite{Dabier_2026}. Recently, state-space models (SSMs) such as Mamba have emerged as efficient alternatives, employed in ProMU \cite{Liu_2025}, MBU-Net \cite{Zhang_2025}, and UNMamba \cite{Chen_2025}. At each stage, new architectures grow progressively larger and more intricate. Currently, UNMamba seems to lead the field by outperforming all state-of-the-art models.

\subsection{Motivation and Contributions}\label{sec:motivation}

This trajectory of increasing complexity raises a natural question: is each added layer of complexity actually responsible for the reported gains? In principle, this question should be straightforward to answer by looking at the results presented in the methods' articles. In practice, it rarely is, because HSU research is affected by a broader reproducibility issue in machine learning \cite{Pineau_2021}: code accompanying published methods is frequently unavailable, incomplete, or dependent on undocumented hyperparameters, preprocessing steps, or initialisation schemes, making independent verification of reported gains difficult or, in many cases, impossible. 

In this context, the contributions of this work are the following:
\begin{itemize}
    \item We introduce a light CNN architecture derived from an ablation study we conducted on UNMamba, a neural network matching SOTA. This new architecture, coined nnAEU, sets a new SOTA, while being an order of magnitude lighter than UNMamba.
    \item We provide a complete, easy-to-implement and reproducible code, tested on several datasets.
\end{itemize}

\section{UNMamba: context and presentation of the ablation study}

\subsection{UNMamba Description}

UNMamba employs a cascaded spatial--spectral Mamba model tailored for blind HSU. By leveraging state-space models (SSMs), it captures long-range spatial and spectral dependencies with linear computational complexity, ensuring scalability. As illustrated in the top part of \autoref{fig:unm-nnaeu}, the model operates in two stages: abundance estimation and endmember estimation. The abundance module first uses 3 UNMamba blocks that extract long-range spatial dependencies with a spatial block, then global spectral features using a spectral block. This forms a unified spatial--spectral representation that is mapped into abundance estimates. Meanwhile, the endmember module incorporates weighted averages of trainable random sequences. Finally, both abundances and endmembers are combined in the LMM module to reconstruct the input HSI. The authors experimentally show that their method obtains highly accurate results.

\subsection{Pruning Procedure}\label{sec:pruning}

To evaluate the role of these complex modules in UNMamba's performance, we perform a simple pruning of the UNMamba blocks used in the abundance module. This leaves the embedding layer, followed by the abundance head. In detail, the embedding module is composed of a 2D convolutional layer reducing the spectral dimension of the input HSI from $B$ to a hidden dimension of 64, a group norm and a SiLU non-linearity. As for the abundance head, it consists of a 2D convolutional layer augmenting the spectral dimension to 128, a group norm, a SiLU, another 2D convolution reducing the spectral dimension to $C$ to reach the abundance size, a batch norm and finally a softmax with scale 3.5. In addition, it uses a dropout layer with probability $5e^{-2}$ after the softmax during training. We keep the endmember and LMM modules untouched, so the pruning procedure only affects the abundance module.
As shown in \autoref{sec:pruning-xp}, the pruned architecture outperforms UNMamba, leading to our new proposed method described in the next section.

\section{Proposed Architecture}\label{sec:nnaeu}

\begin{figure}
    \centering
    \includegraphics[width=\linewidth]{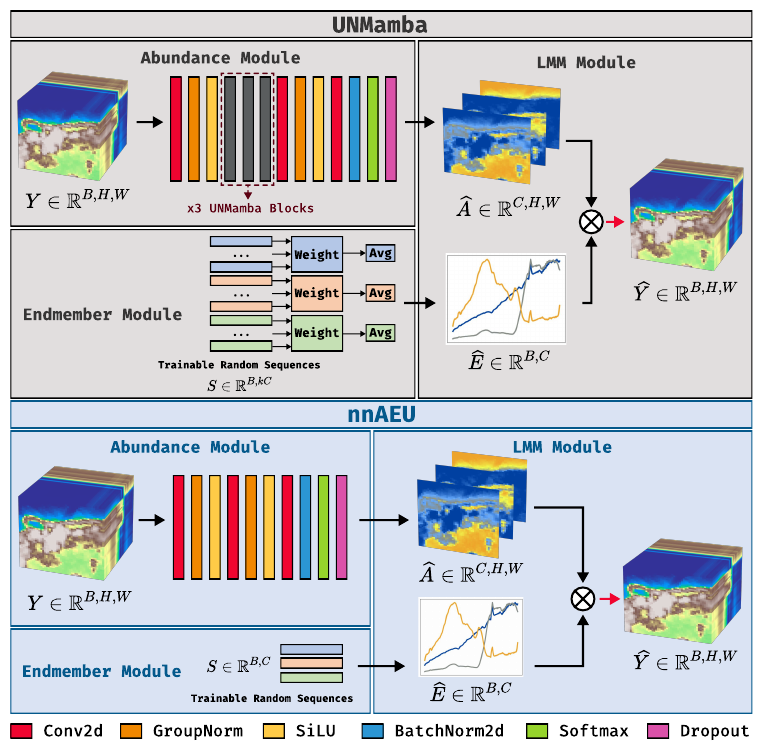}
    \vspace{-20pt}
    \caption{\small{Comparison of UNMamba and proposed nnAEU unmixing networks. nnAEU is derived from UNMamba by pruning the \textbf{UNMamba blocks} and simplifying the endmember module. 
    $Y$ is the input HSI, $\widehat{A}, \widehat{E}$ are the abundance and endmember estimations, $\widehat{Y}$ is the reconstructed HSI and $S$ is the trainable endmember sequence randomly initialised. \textbf{Weight} are trainable multiplicative factors and \textbf{Avg.} represents the averaging operation.}}
    \label{fig:unm-nnaeu}
\end{figure}

The proposed architecture consists of 1) a convolutional network estimating the abundance maps, corresponding to the architecture used in UNMamba but removing the Mamba modules; 2) a linear layer producing the endmembers, simplifying the UNMamba endmember module and further improving its performance. The input HSI is then reconstructed in the LMM module by multiplying both estimated endmembers and abundances. We name this model nnAEU, which stands for Not New Autoencoder Unmixing, as the proposed architecture can be viewed as an autoencoder with the abundance module being the encoder and the endmember estimation as the weights of a decoder. The bottom part of \autoref{fig:unm-nnaeu} illustrates the proposed architecture.

In detail, the encoder uses an alternation of 2D convolutional layers, group norms and SiLU nonlinearities, with the spectral dimension going from $B$ to 48, 128 and finally $C$. It then uses a 2D batch norm, a softmax to enforce both the ANC and ASC and dropout with probability $5e^{-2}$.

In practice, the input HSI is normalised to have values between 0 and 1. Additionally, we use a typical strategy in HSU by initialising the decoder's weights using a first guess of the endmember, using a Simplex Volume Maximisation algorithm \cite{Heylen_2011}.

\subsection{Training loss}

nnAEU is trained end-to-end in an unsupervised manner by minimising a mixture of losses, taken from the UNMamba article :

\vspace{-10pt}
\footnotesize
\begin{equation}
    \text{SAD}(\widehat{Y}, Y) + \lambda_1 \text{MSE}(\widehat{Y}, Y) + \lambda_2 \mathcal{L}_{\frac{1}{2}}(\widehat{A}) + \lambda_3 \displaystyle \sum_{i=0}^C \text{MSE}(\widehat{E}_i, \bar{Y}),
\end{equation}
\normalsize

where $\text{SAD}$ is the Spectral Angular Distance (\ref{eq:sad}), and $\text{MSE}$ is the Mean Squared Error (\ref{eq:mse}), assessing the quality of the reconstruction. In addition, a $\frac{1}{2}$-norm regularisation term encourages sparsity in the abundance estimation (\ref{eq:reg}), and an MSE between the predicted endmembers $\widehat{E}$ and the average spectrum $\bar{Y}$ assists the convergence of the endmembers towards the data distribution, with $\bar{Y} = \frac{1}{N} \sum_{i=0}^{N} Y_{:,i}$. 

Mathematically, for any dumb variable matrices $X,\widehat{X} \in \mathbb{R}^{D_1,D_2}$, we denote:
\vspace{-5pt}
\begin{equation}
    \text{SAD} =  \frac{1}{D_1} \sum_{i=1}^{D_1} \arccos \left( \sum_{j=1}^{D_2} \frac{X_{j,i}}{\| \mathbf{X}_{:,i} \|_2}  \frac{\widehat{X}_{j,i}}{\| \widehat{\mathbf{X}}_{:,i} \|_2} \right),
    \label{eq:sad}
\end{equation}
\vspace{-10pt}
\begin{equation}
    \text{MSE} =  \displaystyle \frac{1}{D_1} \sum_{i=1}^{D_1}(\widehat{X}_{:,i} - X_{:,i})^2,
    \label{eq:mse}
\end{equation}
\vspace{-10pt}
\begin{equation}
    \mathcal{L}_{\frac{1}{2}}(\widehat{A}) = \displaystyle \frac{1}{CN} \sum_{i=0}^{C} \sum_{j=0}^N\sqrt{\widehat{A}_{i,j}}.
    \label{eq:reg}
\end{equation}
\vspace{-20pt}

\section{Experiments}
\subsection{Data Description}

Experiments were conducted on several available labelled datasets, providing an in-depth comparison of the methods :
\begin{itemize}
    \item \textbf{Samson}: the smallest dataset, relatively simple, with a spatial size of $95\times 95$, $156$ spectral bands, and 3 endmembers.
    \vspace{-10pt}
    \item \textbf{Jasper Ridge}: it has $100 \times 100$ pixels, $198$ spectral bands and 4 endmembers.
    \vspace{-10pt}
    \item \textbf{Apex}: has $110 \times 110$ pixels, $285$ spectral bands and contains 4 endmembers.
    \vspace{-10pt}
    \item \textbf{Urban}: has a significantly larger spatial extent of $307$ pixels, 162 spectral bands and 6 endmembers
\end{itemize}

\subsection{Experimental Setup}\label{sec:setup}

The estimated endmember quality is assessed through the SAD (\ref{eq:sad}) and the abundance through the NMSE (\ref{eq:nmse}). Both are averaged over all endmembers. 
\vspace{-5pt}
\begin{equation}
    \text{NMSE} = \displaystyle \frac{\sum_{i=1}^C \sum_{j=1}^N (A_{{\text{GT}_{i,j}}} - \widehat{A}_{i,j})^2}{\sum_{i=1}^C \sum_{j=1}^N (A_{{\text{GT}_{i,j}}})^2}.
    \label{eq:nmse}
\end{equation}
We also compare the execution time in seconds (\emph{Time}), measured on an NVIDIA V100 GPU, as well as the number of parameters (\emph{Size}) for all methods. Additionally, we compute a score $s_i$ for each method $i$, aggregating over all the datasets $j\in \{1..J\}$ (here we consider $J=4$ datasets) the SAD and NMSE after a per-dataset normalisation. We average this score to obtain a global ranking of the methods (\emph{Rank}), for easier readability. The score is obtained with:
\vspace{-5pt}
\begin{equation}
    s_i = \displaystyle  \frac{1}{2J}\sum_{j=1}^{J} \left( \frac{\text{SAD}_{i,j}}{\text{max}_j(\text{SAD}_{i,j})}+\frac{\text{NMSE}_{i,j}}{\text{max}_j(\text{NMSE}_{i,j})}\right), 
\end{equation}
where $\text{SAD}_{i,j}$ and $\text{NMSE}_{i,j}$ are the metrics of method $i$ on dataset $j$. We also include a reproducibility item (\emph{Repr.}) to highlight methods for which announced performance could or couldn't be reproduced.
The training and loss hyperparameters are chosen to be the same across all the datasets. They were obtained using the Optuna Bayesian optimisation framework \cite{optuna_2019}: $300$ training epochs, with the AdamW optimiser, a weight decay of $3e^{-3}$, a learning rate of $7e^{-3}$ and $\lambda_1 = 0.8, \lambda_2=3e^{-2}$ and $\lambda_3=1$.

\subsection{UNMamba Pruning}\label{sec:pruning-xp}

\autoref{tab:pruning} shows results obtained with the ablation described in \autoref{sec:pruning}, where UNMamba-p denotes the pruning of Mamba blocks. This shows that this ablation doesn't affect the performance of the model; in fact, it can even improve unmixing capacity while significantly reducing model size (by a factor of 3 to 5). This size reduction speeds up training and shrinks the memory footprint, making it better suited for embedded systems.

\begin{table*}[h!]
    \centering
    {\fontsize{7pt}{7pt}\selectfont
    \addtolength{\tabcolsep}{-4pt}
    \renewcommand{\arraystretch}{1.3}
    \begin{tabular}{|l|c|c|c|c|c|c|c|c|c|c|c|c|c|c|c|c|c|}
    \hline
    \multirow{2}{*}{} &
    \multicolumn{4}{c|}{\textbf{Samson}} &
    \multicolumn{4}{c|}{\textbf{Jasper}} &
    \multicolumn{4}{c|}{\textbf{Apex}} &
    \multicolumn{4}{c|}{\textbf{Urban}} & \multirow{2}{*}{Score $\downarrow$} \\
    \cline{2-17}
    & \textbf{SAD $\downarrow$} & \textbf{NMSE $\downarrow$} & \textbf{Time} & \textbf{Size} & \textbf{SAD $\downarrow$} & \textbf{NMSE $\downarrow$} & \textbf{Time} & \textbf{Size} & \textbf{SAD $\downarrow$} & \textbf{NMSE $\downarrow$} & \textbf{Time} & \textbf{Size} & \textbf{SAD $\downarrow$} & \textbf{NMSE $\downarrow$} & \textbf{Time} & \textbf{Size} & \\
    \hline
    UNMamba \cite{Chen_2025} & \textcolor{myorange}{\underline{0.025\textsubscript{± 0.01}}} & \textcolor{myblue}{\textbf{0.014\textsubscript{± 0}}} & 32 & 194k & \textcolor{myorange}{\underline{0.095\textsubscript{± 0.03}}} & \textcolor{myorange}{\underline{0.033\textsubscript{± 0.04}}} & 40 & 242k & 0.139\textsubscript{± 0.06} & \textcolor{myorange}{\underline{0.116\textsubscript{± 0.05}}} & 38 & 260k & 0.165\textsubscript{± 0.05} & \textcolor{myblue}{\textbf{0.215\textsubscript{± 0.13}}} & 202 & 3.8M & \textcolor{myorange}{\underline{1.23}}\\
    \hline
    \makecell[l]{UNMamba-p} & \textcolor{myblue}{\textbf{0.024\textsubscript{± 0}}} & \textcolor{myorange}{\underline{0.016\textsubscript{± 0}}} & 9 & 43k & \textcolor{myblue}{\textbf{0.065\textsubscript{± 0.01}}} & \textcolor{myblue}{\textbf{0.019\textsubscript{± 0.01}}} & 12 & 62k & \textcolor{myorange}{\underline{0.122\textsubscript{± 0.02}}} & \textcolor{myblue}{\textbf{0.096\textsubscript{± 0.02}}} & 9 & 85k & \textcolor{myblue}{\textbf{0.148\textsubscript{± 0.04}}} & \textcolor{myorange}{\underline{0.217\textsubscript{± 0.09}}} & 83 & 69k & \textcolor{myblue}{\textbf{0.81}}\\
    \hline
    \end{tabular}
    }
    \caption{\small{Comparison of the performance of UNMamba and UNMamba-p, without the UNMamba blocks. \emph{Time} is in seconds, \emph{Score} is detailed in \autoref{sec:setup}, and \emph{Size} is the number of trainable parameters. Best results in \textcolor{myblue}{\textbf{bold}}, second best \textcolor{myorange}{\underline{underlined}}}}
    \label{tab:pruning}
\end{table*}

\subsection{Results}
\begin{figure*}[h]
    \centering
    \includegraphics[width=\linewidth]{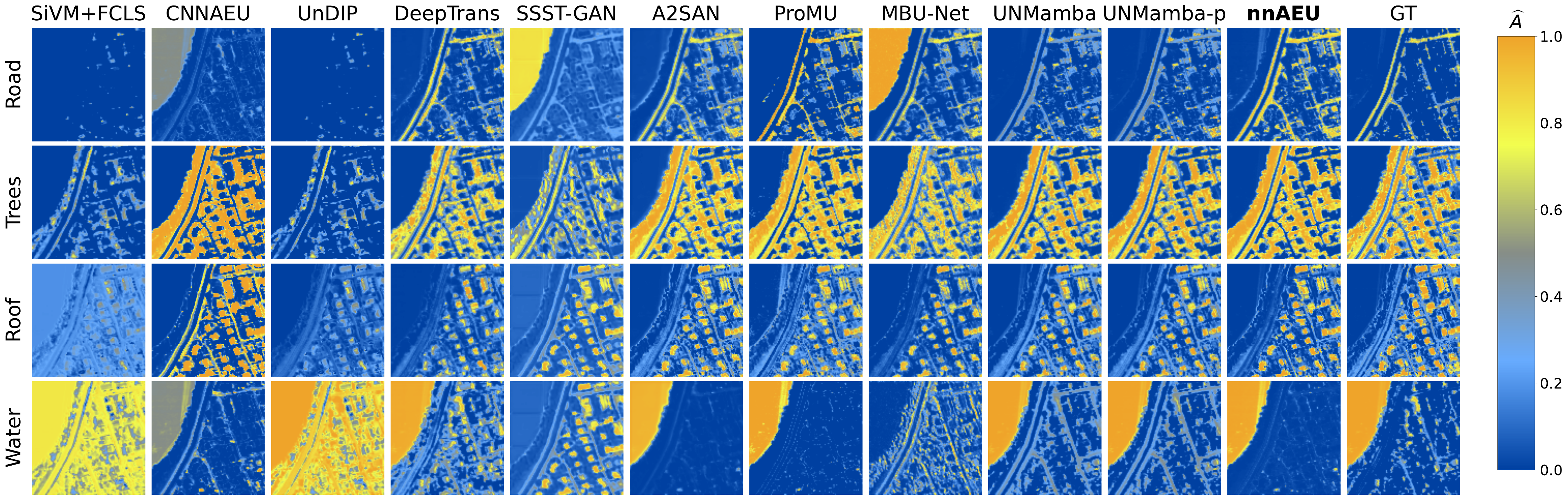}
    \vspace{-20pt}
    \caption{Comparison of all method's estimated $\widehat{A}$ on the Apex dataset, our proposed method is in \textbf{bold}.}
    \label{fig:a-comparison}
\end{figure*}
\begin{figure*}[h]
    \centering
    \includegraphics[width=\linewidth]{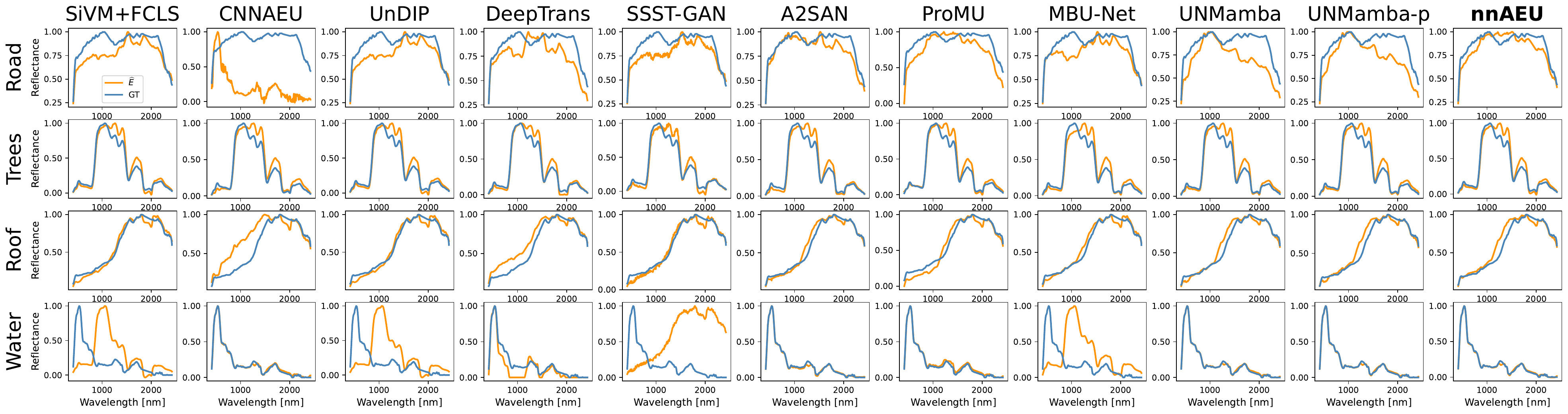}
    \vspace{-20pt}
    \caption{Comparison of all method's estimated $\widehat{E}$ on the Apex dataset, our proposed method is in \textbf{bold}.}
    \label{fig:e-comparison}
\end{figure*}
\begin{table*}[h]
    \centering
    {\fontsize{6.6pt}{7pt}\selectfont
    \addtolength{\tabcolsep}{-4pt}
    \renewcommand{\arraystretch}{1.4}
    \begin{tabular}{|l|c|c|c|c|c|c|c|c|c|c|c|c|c|c|c|c|c|c|}
    \hline
    \multirow{2}{*}{} &
    \multicolumn{4}{c|}{\textbf{Samson}} &
    \multicolumn{4}{c|}{\textbf{Jasper}} &
    \multicolumn{4}{c|}{\textbf{Apex}} &
    \multicolumn{4}{c|}{\textbf{Urban}} & \multirow{2}{*}{\textbf{Repr.}} & \multirow{2}{*}{\textbf{Rank}}\\
    \cline{2-17}
    & \textbf{SAD $\downarrow$} & \textbf{NMSE $\downarrow$} & \textbf{Time} & \textbf{Size} & \textbf{SAD $\downarrow$} & \textbf{NMSE $\downarrow$} & \textbf{Time} & \textbf{Size} & \textbf{SAD $\downarrow$} & \textbf{NMSE $\downarrow$} & \textbf{Time} & \textbf{Size} & \textbf{SAD $\downarrow$} & \textbf{NMSE $\downarrow$} & \textbf{Time} & \textbf{Size} & & \\
    \Xhline{4\arrayrulewidth} 
    SiVM + FCLS \cite{Heinz_1999} & 0.056\textsubscript{± 0} & 0.438\textsubscript{± 0} & 0 & 0 & 0.261\textsubscript{± 0} & 0.187\textsubscript{± 0} & 0 & 0 & 0.244\textsubscript{± 0} & 0.943\textsubscript{± 0} & 0.1 & 0 & 0.436\textsubscript{± 0} & 1.293\textsubscript{± 0} & 0.1 & 0 & \ding{52} & 7 \\
    \hline
    CNNAEU \cite{Palsson_2021} & 0.058\textsubscript{± 0.08} & 0.112\textsubscript{± 0.11} & 1.4 & 124k & 0.253\textsubscript{± 0.17} & 0.328\textsubscript{± 0.16} & 1.4 & 182k & 0.362\textsubscript{± 0.24} & 0.263\textsubscript{± 0.12} & 1.5 & 261k & 0.238\textsubscript{± 0.07} & 0.712\textsubscript{± 0.24} & 5.4 & 188k & \ding{55} & 5 \\
    \hline
    UnDIP \cite{Rasti_2022} & 0.056\textsubscript{± 0} & 0.453\textsubscript{± 0} & 57 & 1.6M & 0.261\textsubscript{± 0} & 0.191\textsubscript{± 0} & 72 & 1.7M & 0.244\textsubscript{± 0} & 1.003\textsubscript{± 0} & 108 & 1.9M & 0.436\textsubscript{± 0} & 1.242\textsubscript{± 0.01} & 269 & 1.6M & \ding{55} & 8 \\
    \hline
    DeepTrans \cite{Ghosh_2022} & 0.198\textsubscript{± 0.01} & 0.346\textsubscript{± 0} & 3.2 & 5M & 0.282\textsubscript{± 0.01} & 0.255\textsubscript{± 0.01} & 3.2 & 7.5M & 0.289\textsubscript{± 0} & 0.445\textsubscript{± 0.01} & 3 & 8M & 0.369\textsubscript{± 0.03} & 0.786\textsubscript{± 0.11} & 9 & 34M & \ding{55} & 10 \\
    \hline
    SSST-GAN \cite{Zhang_2026} & 0.136\textsubscript{± 0.01} & 0.078\textsubscript{± 0.03} & 173 & 175M & 0.304\textsubscript{± 0.06} & 0.259\textsubscript{± 0.04} & 172 & 175M & 0.380\textsubscript{± 0.09} & 0.808\textsubscript{± 0.03} & 179 & 175M & 0.334\textsubscript{± 0.09} & 0.617\textsubscript{± 0.17} & 175 & 175M & \ding{52} & 9 \\
    \hline
    A2SAN \cite{Tao_2024} & 0.060\textsubscript{± 0.01} & \textcolor{myblue}{\textbf{0.014\textsubscript{± 0}}} & 14 & 37k & 0.136\textsubscript{± 0.03} & 0.089\textsubscript{± 0.05} & 14 & 43k & \textcolor{myorange}{\underline{0.080\textsubscript{± 0.01}}} & \textcolor{myorange}{\underline{0.117\textsubscript{± 0.01}}} & 14 & 56k & 0.221\textsubscript{± 0.06} & 0.409\textsubscript{± 0.17} & 16 & 38k & \ding{52} & 3 \\
    \hline
    ProMU \cite{Liu_2025} & 0.106\textsubscript{± 0} & 0.018\textsubscript{± 0} & 431 & 52k & 0.278\textsubscript{± 0.01} & 0.364\textsubscript{± 0.03} & 252 & 17k & 0.146\textsubscript{± 0.05} & 0.188\textsubscript{± 0.10} & 582 & 18k & // & // & // & // & \ding{55} & 6 \\
    \hline
    MBU-Net \cite{Zhang_2025} & 0.028\textsubscript{± 0} & 0.052\textsubscript{± 0} & 144 & 199k & 0.144\textsubscript{± 0.01} & 0.045\textsubscript{± 0.01} & 101 & 144k & 0.293\textsubscript{± 0.04} & 0.941\textsubscript{± 0.26} & 145 & 194k & // & // & // & // & \ding{55} & 4 \\
    \hline
    UNMamba \cite{Chen_2025} & \textcolor{myblue}{\textbf{0.025\textsubscript{± 0.01}}} & \textcolor{myblue}{\textbf{0.014\textsubscript{± 0}}} & 32 & 194k & \textcolor{myorange}{\underline{0.095\textsubscript{± 0.03}}} & \textcolor{myorange}{\underline{0.033\textsubscript{± 0.04}}} & 40 & 242k & 0.139\textsubscript{± 0.06} & \textcolor{myblue}{\textbf{0.116\textsubscript{± 0.05}}} & 38 & 260k & \textcolor{myorange}{\underline{0.165\textsubscript{± 0.05}}} & \textcolor{myorange}{\underline{0.215\textsubscript{± 0.13}}} & 202 & 3.8M & \ding{52} &  \textcolor{myorange}{\underline{2}} \\
    \Xhline{4\arrayrulewidth}
    \textbf{nnAEU (ours)} & \textcolor{myorange}{\underline{0.027\textsubscript{± 0}}} & \textcolor{myorange}{\underline{0.015\textsubscript{± 0}}} & 2 & 15k & \textcolor{myblue}{\textbf{0.066\textsubscript{± 0.02}}} & \textcolor{myblue}{\textbf{0.032\textsubscript{± 0.01}}} & 2 & 17k & \textcolor{myblue}{\textbf{0.064\textsubscript{± 0}}} & 0.127\textsubscript{± 0.01} & 2 & 22k & \textcolor{myblue}{\textbf{0.111\textsubscript{± 0}}} & \textcolor{myblue}{\textbf{0.149\textsubscript{± 0.03}}} & 4 & 16k & \ding{52} & \textcolor{myblue}{\textbf{1}} \\
    \hline
    \end{tabular}
    }
    \caption{Comparison of all HSU methods on the 4 datasets, averaged over 10 runs.  We display results as Mean\textsubscript{± std}, \emph{Time} (in seconds), \emph{Rank} (as described in \ref{sec:setup}), \emph{Size} (number of trainable parameters), and \emph{Repr.} indicates whether the method's announced performance could be replicated or not.}
    \label{tab:methods-comparison}
\end{table*}

To evaluate our proposed framework, we compare averaged scores from 10 independent training runs against 9 state-of-the-art methods. These methods span a diverse range of approaches, from classic optimisation method using SiVM and FCLS \cite{Heylen_2011, Heinz_1999}, to CNNs with CNNAEU \cite{Palsson_2021} and UnDIP \cite{Rasti_2022}, ViTs with DeepTrans \cite{Rasti_2022}, A2SAN \cite{Tao_2024}, GAN with SSST-GAN \cite{Zhang_2026}, and Mamba with ProMU \cite{Liu_2025}, MBU-Net \cite{Zhang_2025} \footnote{we could not get results on the Urban data set for ProMU using the available code due to missmatch in parameters, and for MBU-Net due to exceeding memory issues}, and UNMamba \cite{Chen_2025}. Qualitative results about the estimation of abundances and endmembers on the Apex dataset are shown in Figures \ref{fig:a-comparison} and \ref{fig:e-comparison}. In addition, \autoref{tab:methods-comparison} displays the results obtained using the proposed unmixing autoencoder. Moreover, as mentioned in \autoref{sec:motivation}, it is important to note that some methods' results may differ from their original paper due to reproducibility issues with the available code.

We observe that, as mentioned in \autoref{sec:nnaeu}, our proposed simplification of the endmember module further improves the performance compared to using UNMamba's (\autoref{tab:pruning}). This choice reduces the size of the already pruned UNMamba model, reaching up to 20 times fewer parameters than UNMamba, placing it among the smallest and fastest learning-based models of the state-of-the-art. In addition, this method sets a new state-of-the-art, beating all other approaches to unmixing, surpassing the pruned version of UNMamba. 

Our findings should be read with two limitations in mind: our evaluation is restricted to standard benchmark scenes, itself a consequence of HSU's scarce, costly-to-obtain ground truth; and Mamba- or transformer-based components may still prove useful in other regimes — larger datasets, more endmembers, different sensors — where more data could justify their added capacity.
\vspace{-10pt}
\section{Conclusion}
Why doesn't added complexity improve results? A likely explanation lies in the single-image training framework, dominant in HSU: heavy, expressive models are trained on a single scene with little data, leaving limited room for their extra capacity to be exploited — the convolutional backbone alone may already capture what signal is available.

Within current benchmark conditions, our results support a clear message: complexity should not be treated as a proxy for progress. 
We demonstrated this through a controlled ablation of UNMamba, and proposed a light, fully reproducible CNN architecture matching SOTA at an order of magnitude fewer parameters, offered as a simple baseline for the field.


\bibliographystyle{IEEEbib}
\bibliography{strings,refs}

\end{document}